\documentclass[letterpaper, 10 pt, conference]{ieeeconf}  
\IEEEoverridecommandlockouts                              
                                              
\usepackage{float}
\usepackage[normalem]{ulem}
\usepackage{amsmath}
\usepackage[noadjust]{cite}
\usepackage{siunitx}
\usepackage{amssymb}

\makeatletter
\let\NAT@parse\undefined
\makeatother
\usepackage[colorlinks,bookmarksopen,bookmarksnumbered,citecolor=red,urlcolor=red]{hyperref}
\usepackage[capitalise]{cleveref}

\title{\LARGE \bf
Function-based Parametric Co-Design Optimization of Dexterous Hands}

\usepackage[colorinlistoftodos]{todonotes}

\usepackage{comment}

\usepackage{booktabs}
\usepackage{tabularx}
\usepackage{array}

\begin{document}


\author{Mohammad Amin Mirzaee$^{1}$,
        Harsh Gupta$^{1}$,
        and Wenzhen Yuan$^{1}$
        \\ \href{https://www.aminmirzaee.com/HandCDO/}{\nolinkurl{aminmirzaee.com/HandCDO/}}\vspace{-10pt}
\thanks{$^{1}$ are with University of Illinois at Urbana-Champaign, Champaign, IL, USA 
        {\tt\small \{mirzaee2,hgupt3,yuanwz\}@illinois.edu}}%
        }

\maketitle

\thispagestyle{empty}
\pagestyle{empty}

\begin{abstract}
Despite advances in dexterous hand manipulation, robotic hand design is still largely decoupled from task-driven evaluation and control, limiting systematic optimization. Existing robotic hand co-design approaches are often limited in scope, optimizing a small subset of design parameters. We introduce a comprehensive parametric framework for robotic hand generation that unifies palm structure, finger kinematics, fingertip geometry, and fine-scale surface curvatures within a single design space. Fine geometric features are introduced through parametric surface deformation kernels that directly influence contact interactions.
We validate the framework on design optimization in grasp stability tasks in simulation and real-world dynamic scenarios. 
Our framework produces simulation- and fabrication-ready hand models and will be released as open-source to enable rapid design iteration for dexterous hand co-design optimization frameworks and cross-embodiment policy training and control research.
\end{abstract}

\IEEEpeerreviewmaketitle

\section{Introduction}

What happens if we lose a finger or two on our hands? What if we had more joints on the fingers? Or if the shape and length of them were different. Would that affect our performance? Intuitively, we know they do alter our ability to interact with the physical world and our ultimate dexterity after adaptation to our new hands. It is not just about time and remapping the brain to them; it is what we are remapping to.
Another important question is whether our hands are the best for a specific task. 
The answer is beyond sensing and control; it concerns the kinematics, geometry, and actuation.

We have achieved versatility and dexterity through a tight coupling of morphology, sensing, and control over years of evolution \cite{gilday2023sensing, marzke2000evolution}. It is the harmony of these features that allows our hands to conform to complex geometries, distribute forces intelligently, and adapt instantly to uncertainty. The shape of the hand determines where contacts can form, sensing reveals how those contacts behave, and control coordinates the response. Dexterity emerges not from any single component but from their integration.

Unlike human hands that have evolved in different aspects, most robotic hands and systems are designed using a sequential pipeline in which hardware is manually designed and fixed prior to control optimization \cite{christoph2025orca, shaw2023leap}. This separation results in designs that limit manipulation performance and require controllers to compensate for suboptimal embodiment \cite{calandra2018more, yin2025geometric}. However, control policy performance is highly sensitive to hand geometry, actuation, and contact properties. These observations motivate co-design optimization (CDO), in which hand morphology and control are optimized jointly with respect to task objectives \cite{cheney2014unshackling, bai2025learning, pathak2019learning}. Co-design enables the discovery of design–control combinations that are difficult to obtain through manual tuning.

In a hand co-design optimization, we generate new hand designs and configurations (similar to mutation in biology) to be evaluated using specific metrics, such as optimal control policy performance (neural adaptation in biology). Then, we need to loop this process (evolution) in an optimizer to tune hand configurations and designs. Through iterations, the optimal design is achieved \cite{deimel2017automated, fay2025cross}.

\begin{figure}
    \centering
    \includegraphics[width=\linewidth]{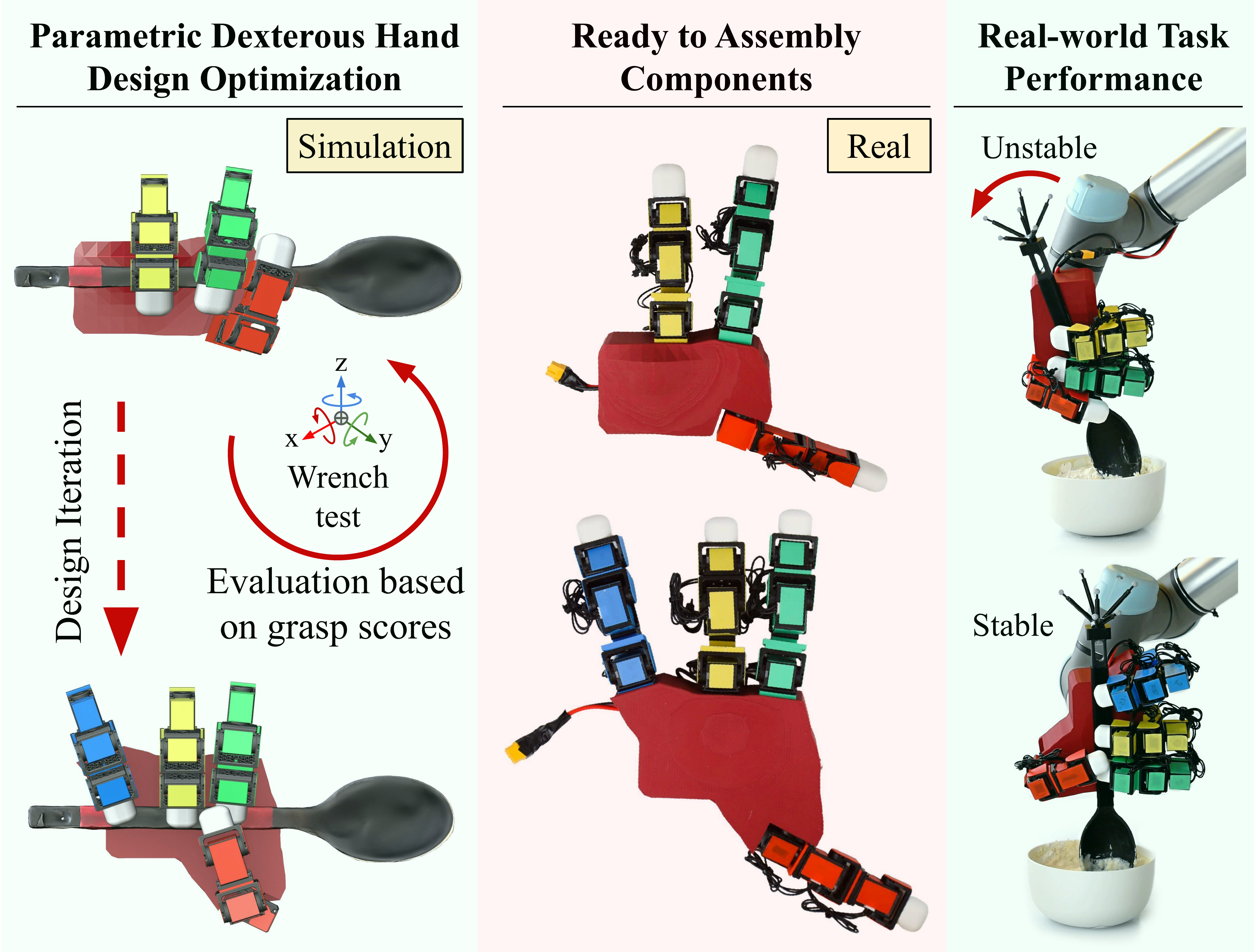}
    \vspace{-18pt}
    \caption{Our hand co-design optimization framework finds the best hand design and grasp configurations for a specific task. Our model outputs ready-to-fabricate and assemble components for real-world task scenarios.}
    \label{fig:placeholder}
    \vspace{-15pt}
\end{figure}

In the generation stage, we require a model capable of producing and modifying diverse hand designs, such as adding fingers or joints, adjusting offsets, and reshaping the palm. This demands explicit parameterization of the high-dimensional design space spanning kinematics and geometry. Prior work typically explores only limited parameter subsets \cite{yi2025co, mannam2023framework, deimel2017automated, meixner2019automated, fay2025cross, mannam2024design}, resulting in narrowly scoped frameworks. To address this gap, we propose a comprehensive hand design parameterization and generation system that spans from high-level palm and finger configurations to fine-grained surface and fingertip geometries, enabling task-dependent prioritization of design variables.

Our hand generator outputs a configured dexterous manipulator URDF and corresponding meshes, including accurate colliders and assembly-aware visualization bodies, which can be imported into simulators or directly printed and assembled, bridging the sim-to-real hardware gap that exists in many co-design optimization frameworks \cite{vaish2024co, you2019kinematic, deimel2017automated}.

After generating new hand designs, we evaluate them by assigning a simulation-based performance score for iterative optimization. We focus on grasp stability. From picking up an object or holding a tool to transitioning between grasps in an in-hand manipulation task, we require stability and robustness against disturbances and external forces like gravity \cite{weisz2012pose, miller2004graspit, wang2022dexgraspnet}. Specifically, we optimize for stable power grasps of tools (e.g., a hammer), using an inner optimization loop to compute the best grasp configuration for each generated hand. While grasp stability is our target metric, the framework readily extends to other tasks with custom performance objectives.

Optimization closes the loop between hand generation and evaluation by proposing new design configurations for simulation. Because common simulation-based dexterous grasp evaluation involves multi-contact interactions in non-differentiable physics simulators such as Isaac Sim and MuJoCo, and includes discrete and categorical design parameters, gradient-free optimization methods are preferred \cite{deimel2017automated, luck2020data}. These approaches do not require derivative information or smooth objective functions and are well-suited for complex, non-convex, discontinuous, and noisy systems like ours \cite{rios2013derivative, larson2019derivative}.

To validate our framework, we conduct a co-design optimization over 28 kinematic and surface-level geometric parameters, including joint placement, link dimensions, fingertip scaling, and palm features. Designs are evaluated using a grasp stability metric derived from simulated optimal grasps of three tools. We perform SHAP analysis to quantify each parameter’s contribution to performance. We further validate the results through real-world dynamic grasping experiments using three fabricated hands.

Although demonstrated on a subset of parameters for a multi-fingered hand in a power grasp task, the framework is not limited to a single embodiment or scenario. The key contribution lies in formalizing almost every aspect of a hand design as a parameterized and searchable space, where morphology becomes a programmable variable rather than a fixed constraint. By redefining the design space for a new hand, through alternative kinematic structures, surface geometries, or task-specific constraints, the same pipeline can be directly applied to optimize different robotic hands for different objectives.

\section{Related Works}

\subsection{Robot Hand Design Parameterization}

Robotic hand design has traditionally followed two directions: anthropomorphic replicas of the human hand and simplified grippers for specific industrial tasks. Many systems adopt fixed bio-inspired architectures that replicate anatomical features such as skeletal structure and tendon routing \cite{grebenstein2012hand}, with platforms like the Shadow Hand and Allegro Hand exemplifying this approach. However, these embodiments are typically handcrafted and remain static, with limited task-driven optimization.

To move beyond manual design, several works introduced explicit parameterizations of hand morphology. Early studies explored constrained mechanical subspaces, such as optimizing pulley routing in underactuated hands \cite{cabas2006optimized} or tuning geometric parameters to improve grasp success \cite{dong2018geometric}. Later efforts expanded the search space to structural variables like finger count, placement, and joint number \cite{meixner2019automated}, cage-based link scaling \cite{xu2021end}, and optimization of physical properties such as friction and damping under fixed kinematics \cite{fayhardware}.

Recent work has broadened the parameter space in two complementary directions. On the structural side, Gilday et al. parameterized tendon-driven hands over joint diameter, ligament thickness, and pulley characteristics with hardware validation \cite{gilday2025embodied}, while Fay et al. \cite{fay2025cross} introduced a morphology-conditioned framework for jointly exploring diverse structural configurations. In the soft robotics domain, Yi et al. parameterized flexure-based grippers by block-wise stiffness distribution, using a neural physics surrogate for differentiable end-to-end optimization \cite{yi2025co}, and Bai et al. derived a design space for tendon-driven soft hands spanning finger geometry and actuation routing, paired with a learned reward model for efficient search \cite{bai2025learning}.

On the contact-geometry side, Chen et al. represented fingertip modifications as a simplified contact plane and jointly optimized its parameters with control to enable both power and precision grasps within a single hand \cite{ye2025power}.

Despite these advances, most approaches address either coarse structural variables or narrow contact-level modifications in isolation. A unified parameterization spanning kinematic structure, fine-grained surface geometry, and contact-level features within a single extensible framework remains an open challenge.

\subsection{Robot Hand Co-Design and Optimization}

Optimization of robotic systems has long been studied for joint configurations and control using simplified kinematic or dynamic models \cite{yu1998optimization, mayorga1992kinematic, vijaykumar1985geometric}, but hardware-level design optimization traditionally required significant manual intervention. Recognizing the strong coupling between embodiment and control, recent research has shifted toward co-design frameworks that jointly optimize morphology and policy \cite{ha2017joint, chen2023c, chen2021co, whitman2020modular}, with reinforcement learning further enabling joint optimization in locomotion domains \cite{whitman2023learning, schaff2019jointly}.

For robotic hands, co-design studies have grown increasingly sophisticated but remain limited in scope. Fay et al. presented a cross-embodied co-design framework with a morphology search space paired with morphology-conditioned control policies, enabling an end-to-end pipeline that can design, train, fabricate, and deploy a new hand \cite{fay2025cross}. At the contact level, Chen et al. showed that jointly optimizing fingertip geometry and manipulation policy can substantially expand the functional range of a fixed hand platform \cite{ye2025power}. Yi et al. replaced expensive simulation with a differentiable neural surrogate, enabling gradient-based co-optimization of gripper stiffness and grasp pose with 3D-printed hardware validation \cite{yi2025co}. Bai et al. instead trained a reward model on teleoperation data and combined it with a cross-entropy method, reducing required design evaluations by more than half while learning a distribution of high-performing soft hand designs \cite{bai2025learning}.

Despite this progress, existing methods typically optimize a small subset of parameters and target narrow task objectives. Comprehensive co-optimization across heterogeneous design variables, spanning kinematics, geometry, material properties, and contact features, remains underexplored, particularly for stability-driven, task-conditioned optimization across diverse embodiments and grasp types.

\section{Defining and Modeling Hand Designs}

\begin{figure}
    \centering
    \vspace{3pt}
    \includegraphics[width=\linewidth]{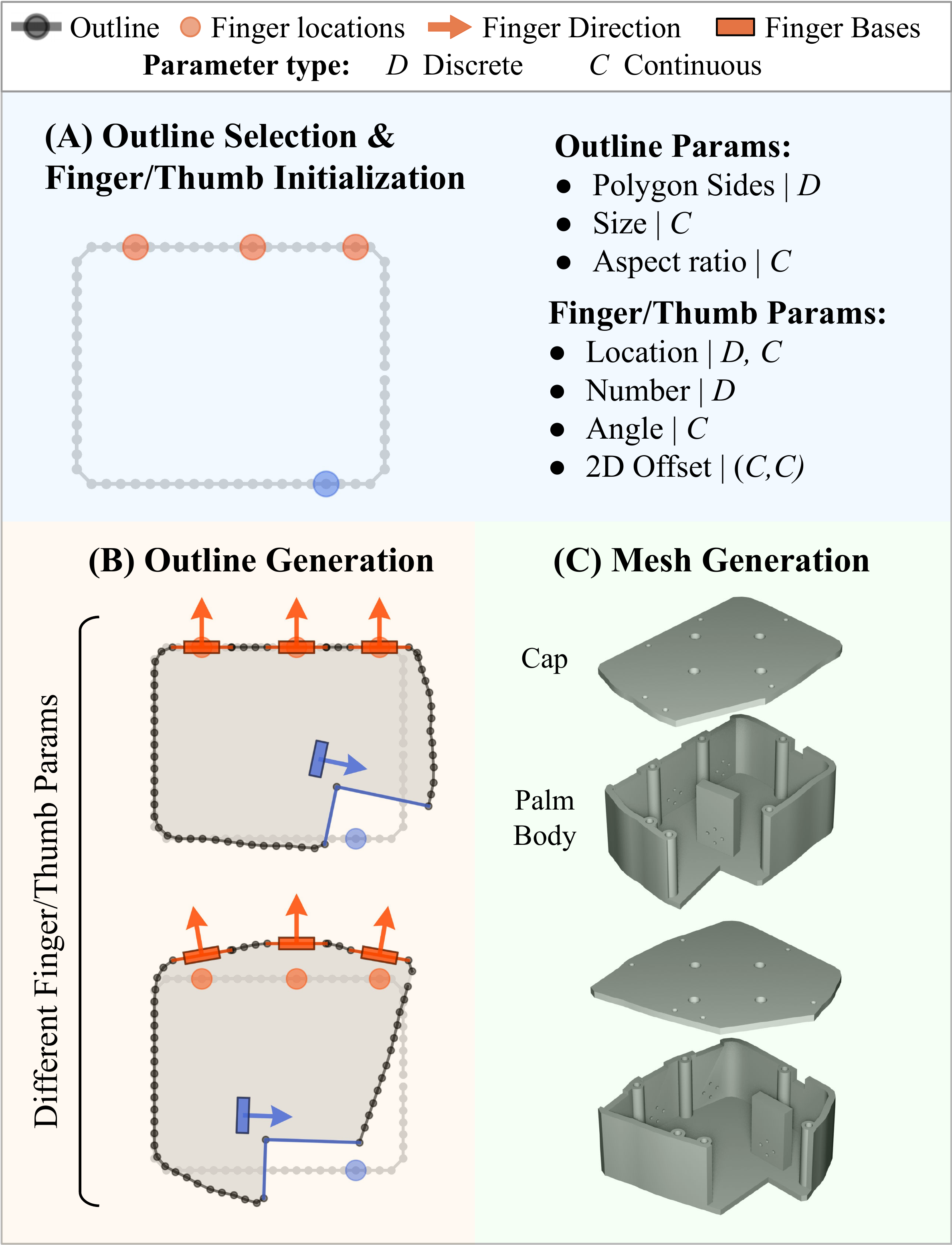}
    \vspace{-18pt}
    \caption{Palm parameterization.
    (A) An initial outline is defined with parameters specifying finger and thumb base locations. Blue colors correspond to the thumb and red colors to the fingers.
    (B) Varying these parameters alters the overall palm shape.
    (C) The resulting outline is converted into assembly-ready palm bodies.}
    \label{fig:Palm}
    \vspace{-10pt}
\end{figure}

\begin{figure*}
    \centering
    \vspace{4pt}
    \includegraphics[width=\linewidth]{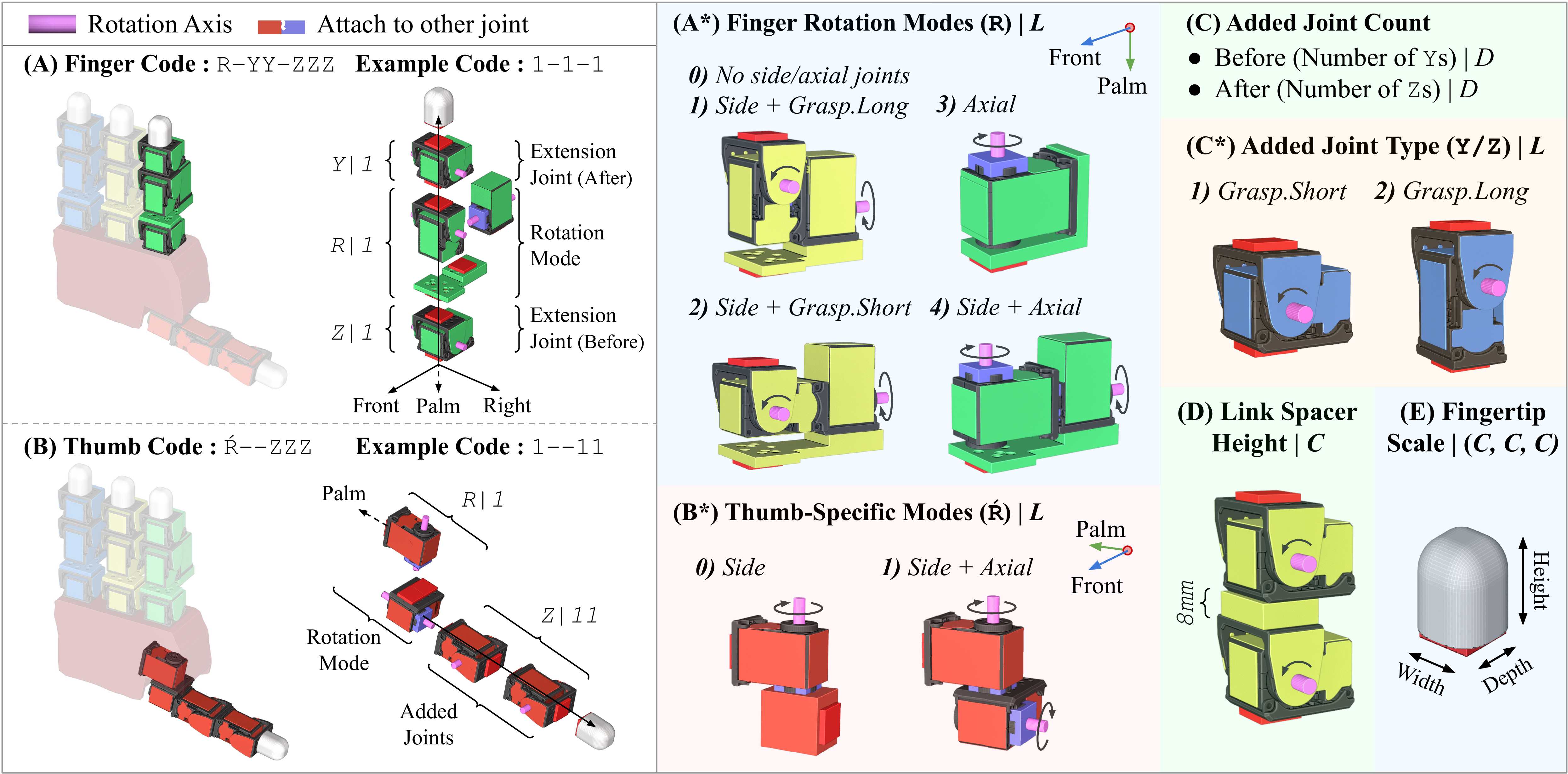}
    \vspace{-18pt}
    \caption{Finger and thumb structural parameterization.
    (A,B) Modular codes define joint base rotation mode and optional additional joints.
    (A*,B*) Different configurations yield distinct degrees of freedom for base rotation.
    (C,D) Additional closing joints extend the base rotation.
    (D,E) Link dimensions and fingertip geometries are parametrically varied.}
    \label{fig:Finger}
    \vspace{-10pt}
\end{figure*}

We generate the hand in two stages. First, we construct the palm geometry and define the finger attachment points based on the design parameters, then structure and assemble each finger onto the palm. In the second stage, we deform the surface geometry of both the palm and fingers.

The hand-generation process is configured by a large set of parameters, including continuous and discrete variables as well as a number of categorical choices. Parameters may be manually specified, randomly sampled, or optimized following an initial exploration phase.

\subsection{High-level Structure}
\subsubsection{Palm Model}
We parameterize the palm geometry as an extrusion of a planar 2D shape. The surface outline is defined by the parameters listed in \cref{fig:Palm}A. Palm size, polygon sides, and aspect ratio determine the initial outline.

The fingers and the thumb are initially assigned to the locations on the outline. Secondary continuous parameters defining the finger base location and orientation (\cref{fig:Palm}A) are then specified. Finally, the palm outline is updated by incorporating the base segments and applying additional integrity steps, \cref{fig:Palm}B, to output the final mesh\cref{fig:Palm}C.

\subsubsection{Finger Model}
Each finger is attached to its corresponding base on the palm. The finger structure is defined by a set of parameters and represented by a code, as shown in \cref{fig:Finger}A. The first parameter in the code, rotation mode (R), diversifies the enabled joint types, from a single-axis Grasp joint ($R=0$, revolves around the right axis) to a three-axis joint selection ($R=4$, side and axial joints with an optional Grasp joint). The added joint count and types \cref{fig:Finger}C and C*, whether they attach before or after the rotation mode joints, introduce several new parameters to the finger configuration.

For the thumb, we also introduce a specialized joint configuration inspired by the Leap Hand \cite{shaw2023leap}, which includes an initial lateral joint at the base. This ensures a minimum level of dexterity and enables opposability comparable to that of the human thumb. Two thumb-specific rotation modes determine whether an axial joint connects to the first joint or not, \cref{fig:Finger}B*.

We modify the existing link lengths through the link spacer height parameter group, \cref{fig:Finger}D, as well as modify the fingertip size in all three axes using scaling parameters.

Through this parameterization, we can generate a great number of diverse finger kinematics that structure robot hands with different levels of dexterity. Each of these finger structures might outperform others in a specific task, which can be determined in the evaluation and optimization stage. For example, a finger with an axial joint connected to the fingertip can potentially rotate the objects in hand without breaking contact, which could outperform a hand that only possesses closing joints.

\subsubsection{Fine Geometry Modeling}

\begin{figure}
    \centering
    \vspace{5pt}
    \includegraphics[width=\linewidth]{figs/SurfaceDeform_v2.pdf}
    \vspace{-18pt}
    \caption{Palm surface geometry parameterization via surface kernels. (A) Parametric Gaussian kernels. (B) We use multiple kernels and superimpose them to produce a curved but smooth surface for the palms.}
    \label{fig:surfacedeform}
    \vspace{-10pt}
\end{figure}

Surface curvatures significantly influence performance in both static and dynamic manipulation tasks by altering contact reaction forces and grasp stability. For example, a concave palm can potentially improve the grasping of tools such as a hammer compared to a flat surface.

We generate surface pads on the base meshes of the hand components and deform their top surfaces, as shown in \cref{fig:surfacedeform}. Deformations are applied using parametric surface kernels (e.g., Gaussian), which displace vertices along their surface normals within selected center regions. Each kernel is controlled by a set of parameters illustrated in \cref{fig:surfacedeform}A.

We decompose the pad mesh into small colliders, \cref{fig:surfacedeform}B, for compatibility with most of the physics simulators that only support convex mesh import from the URDF.

\subsection{Hand Generation}
Once all parameters are specified, the hand is generated and exported as a complete URDF, including all associated collision geometries with precise modeling, ready to be imported into physics simulators as an articulated robot model with relevant kinematic and inertia information. The resulting STL files require no additional processing and are ready for fabrication and assembly.

\begin{figure}
    \centering
    \vspace{5pt}
    \includegraphics[width=\linewidth]{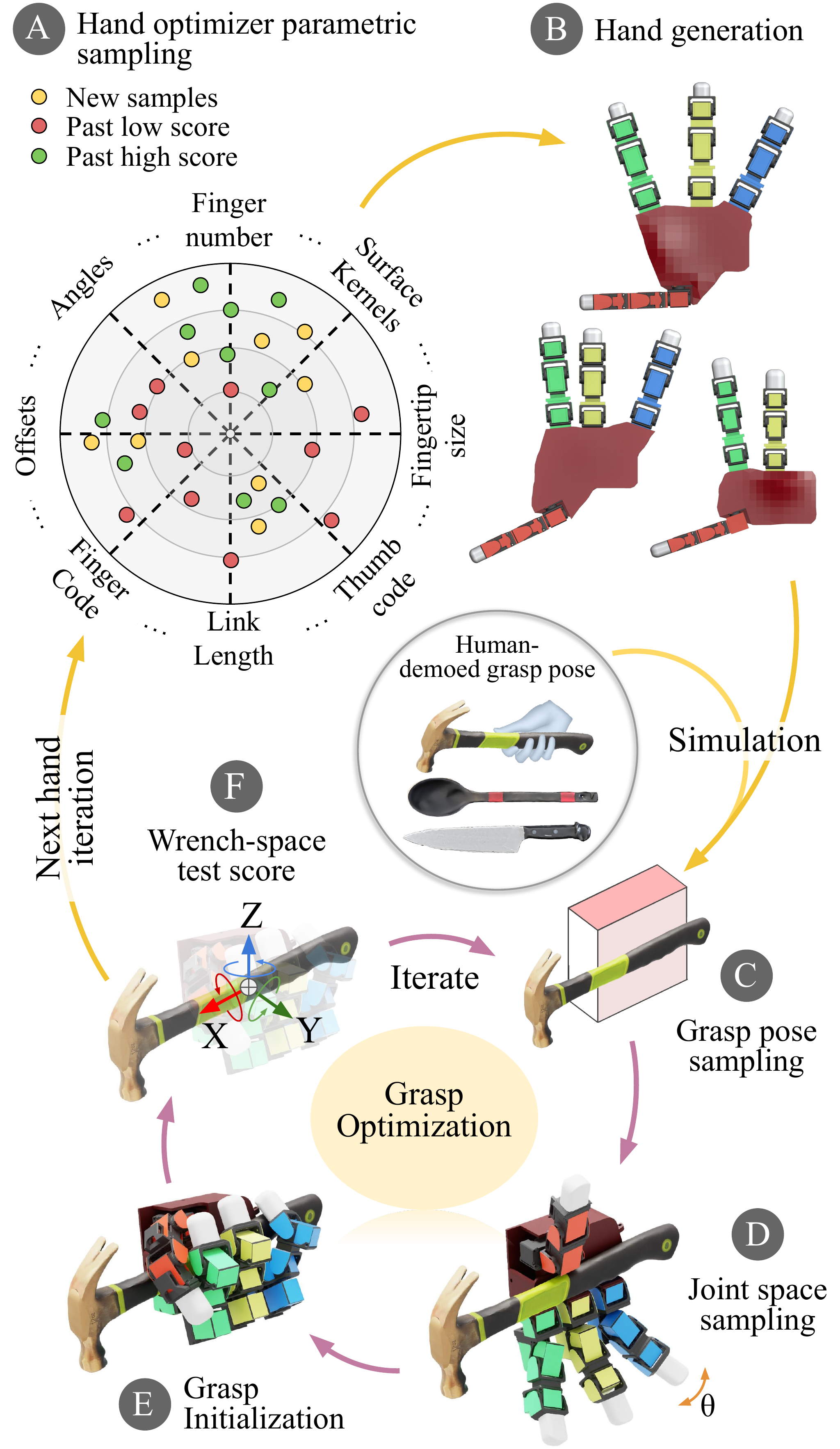}
    \vspace{-18pt}
    \caption{Optimization framework overview.
    (A-B) The hand optimizer samples new hand designs from predefined parameter ranges and custom structural specifications.
    (C-F) We build upon the grasp synthesis and wrench-space evaluation method of \cite{gupta2026grasp}, extending it with an optimization loop over grasp configurations to identify the most stable grasps for each hand-tool pair.
    For each generated hand, grasp optimization is performed across the target tool set. The resulting performance scores are then fed back to the hand optimizer to guide subsequent design iterations.}
    \label{fig:framework}
    \vspace{-10pt}
\end{figure}

\section{Task-driven Design Evaluation}

To demonstrate the hand generation and optimization pipeline, we optimize a subset of design parameters for power grasp tasks in simulation (Isaac Sim 4.5). We adapt the method presented in \cite{gupta2026grasp} with a few improvements. 

Unlike the reference method that iterates through new grasps in a purely random search manner, we synthesize grasp candidates through systematic optimization and select the most stable one for each of the tools. This structured search not only improves sample efficiency but also ensures that each morphology is evaluated under its best achievable grasp strategy, enabling a fair and task-driven comparison across different hand designs. The evaluation method measures grasp stability by testing robustness to applied disturbance wrenches on the tool.

\paragraph{Grasp Generation}
A grasp configuration is defined as $G = \{\mathbf{q}_0,T_{\text{grasp}}\}$,
where $\mathbf{q}_0 \in \mathbb{R}^n$ denotes the finger joint angles and $T_{\text{grasp}} \in SE(3)$ is the wrist pose expressed in the object frame. During grasp synthesis, the object is initially suspended.
Candidate wrist poses, as the first set of optimization parameters, are sampled within a functional region, \cref{fig:framework}C, centered on a reference human grasp wrist pose $T^{\tau}_{\text{wrist}}$ demonstrated for each tool, according to: $T_{\text{grasp}} = T^{\tau}_{\text{wrist}} \cdot T_{\text{perturb}}$
where $T_{\text{perturb}}$ consists of bounded translational and rotational perturbations.

Finger configurations are diversified by sampling inter-finger spread angles for individual fingers, a second set of parameters. We fix all the joint torque thresholds that terminate finger motion after contact at a much lower threshold, 20 percent of the minimum value for the reference method \cite{gupta2026grasp}, to suit our optimization framework without early-stage score saturation. The current sampling strategy enables the synthesis of a wide range of grasp configurations that adapt to tools with varying shapes and orientations.

\paragraph{Grasp Stability Scoring}
Each candidate's grasp configuration $G$ is evaluated using disturbance-based wrench tests applied in Cartesian wrench space. External forces and torques are independently applied along all positive and negative Cartesian axes, yielding 12 test directions. For each direction $i$, the wrench magnitude is increased linearly up to a maximum force $F_{\max}$ or torque $\tau_{\max}$ over a fixed duration $t_{\max}$. Grasp failure is detected when object motion exceeds position or orientation thresholds $(\delta_p, \delta_\theta)$, at which point the stable duration $t^{(i)}_{\text{stable}}$ is recorded.
The grasp stability score is computed as the average normalized stable time across all directions for a specific tool $t$:
\begin{equation}
S_t = \frac{1}{12} \sum_{i=1}^{12} \frac{t^{(i)}_{\text{stable}}}{t_{\max}}.
\label{eq:graspscore}
\end{equation}

For each hand-tool combination, we find the optimal grasp to form the final hand score $S_h$ by taking the average of the $K$ best grasp scores for $T$ tools \cref{eq:handscore}. In the current paper, we use three tools: a hammer, a spoon, and a knife.

\begin{equation}
S_h = \frac{1}{K \times T} 
\sum_{t=1}^{T}  \sum_{k=1}^{K} S^k_t
\label{eq:handscore}
\end{equation}

\paragraph{Dynamic task-specific instability}
After finding the optimal hand and grasp, we analyze the grasp stability for three hands performing tasks on three tools in an open-loop real-world dynamic setting following the human-demonstrated trajectory of hammering a nail, cutting a cucumber, and mixing a pancake batter. We use an OptiTrack motion-capture system with a rigid mount attached to each tool to accurately
record its pose during real-world trials. We then report the relative in-hand rotation of the tools by comparing it to the robot's recorded end effector pose.

\section{Design optimization}

Candidate hand designs are evaluated through physics-based simulation, outputting a non-differentiable performance score. Also, the design space contains a mixture of continuous, discrete, and categorical parameters, as well as conditional dependencies between parameters. These characteristics prevent the use of gradient-based methods.

\subsection{General Problem Formulation}
Let $\mathbf{x} \in \mathcal{X}$ denote a hand design parameter vector, where $\mathcal{X}$ is a mixed search space consisting of continuous parameters (e.g., link lengths, offset values), discrete parameters (e.g., joint counts, number of fingers), and categorical parameters (e.g., rotation modes). The objective is to maximize a task-dependent performance metric from simulation:
\begin{equation}
\mathbf{x}^* = \arg\max_{\mathbf{x} \in \mathcal{X}} f(\mathbf{x}),
\end{equation}
where $f(\mathbf{x})$ is computed by running a full simulation rollout of the manipulation task and measuring, for example, task success, stability, and/or efficiency. Due to stochasticity in contact dynamics and control, $f(\mathbf{x})$ may be noisy and is treated as a stochastic black-box function.

\subsection{Bayesian Optimization with TPE}

In this work, we adopt the Tree-structured Parzen Estimator (TPE) as the optimizer in our co-design framework for both grasp optimization and hand optimization. TPE is a sequential model-based optimization method that efficiently searches diverse design spaces composed of continuous, discrete, and categorical variables. Moreover, TPE scales well to moderately high-dimensional search spaces and remains robust under noisy objective evaluations arising from stochastic grasp synthesis and physics-based simulation.

Unlike classical Gaussian Process–based Bayesian optimization, which models the conditional distribution $p(f \mid \mathbf{x})$, TPE models the inverse density $p(\mathbf{x} \mid f)$ using non-parametric density estimators.

At each iteration, the set of previously evaluated designs is partitioned into two subsets based on a performance quantile threshold, $f^*$: a set of high-performing designs, $G$, and a set of lower-performing designs, $B$. TPE fits separate density models to these two subsets, $b(x)$ and $g(x)$, and selects new candidate designs by maximizing the ratio between the likelihood of sampling from the high-performing distribution and the low-performing distribution, \cref{eq:maximize}. This mechanism naturally balances exploration of uncertain regions with exploitation of promising design configurations.

\begin{equation}
    g(\mathbf{x}) = p( \mathbf{x} \mid f(\mathbf{x}) \geq  f^* ), \quad
    b(\mathbf{x}) = p( \mathbf{x} \mid f(\mathbf{x}) < f^* )
\end{equation}

\begin{equation}
    \mathbf{x}_{\text{next}} = \arg\max_{\mathbf{x}} \frac{g(\mathbf{x})}{b(x)}
    \label{eq:maximize}
\end{equation}

We use separate TP-based optimizers with a similar structure for both grasp optimization and hand optimization loops. The only difference is the set of optimization parameters and the objective function to maximize, \cref{eq:graspscore,eq:handscore}.

\begin{figure*}
    \centering
    \vspace{4pt}
    \includegraphics[width=\linewidth]{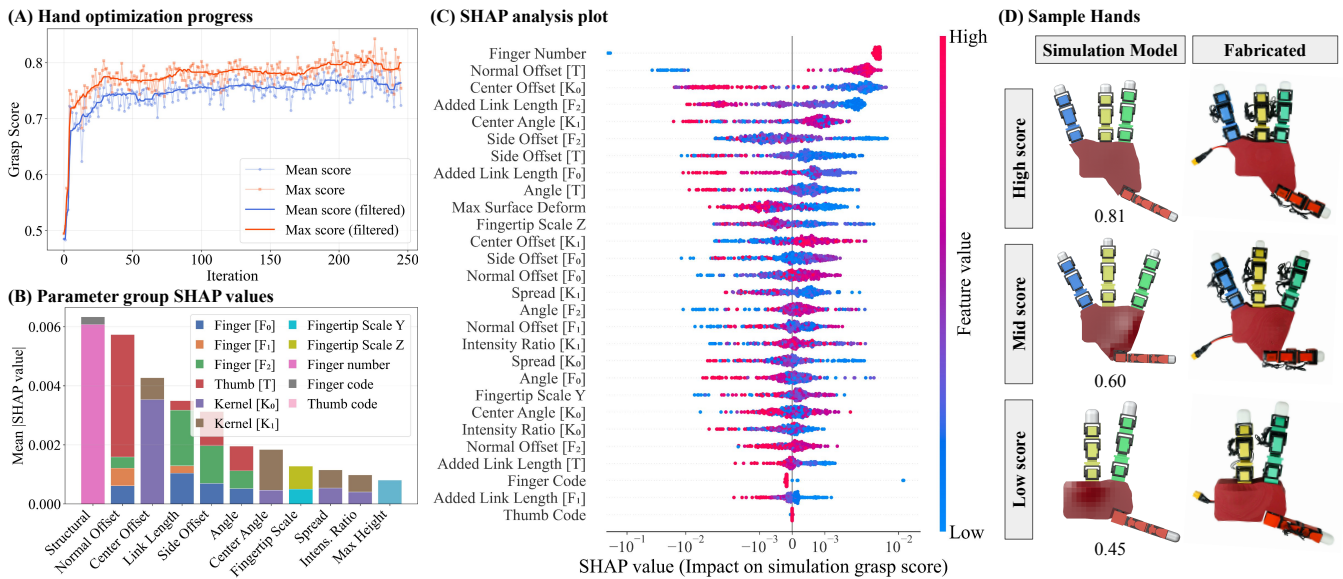}
    \vspace{-18pt}
    \caption{Hand-design optimization results in simulation and parametric analysis. (A) The optimization converges after 200 iterations. After the first exploration phase, the two-finger hands are not selected due to consistently low grasp scores. (B, C) SHAP analysis for ranking and investigating the effect of each design parameter on the final grasp score. B shows the importance of the parameter groups, while C illustrates how individual parameters vary the grasp score. Blue and red colors correspond to the parameter range. (D) We choose three hands of different scores and fabricate them for experiments.}

    \label{fig:simres}
    \vspace{-10pt}
\end{figure*}

\section{Experiments}
We perform optimization of 28 hand parameters selected from different design categories, listed in \cref{tab:optimization_parameters}. We customize the parameter ranges intuitively based on the tool sizes and compatibility with the simulation task. The palm pad resolution is set to a moderate value to accelerate the process. The grasp configuration ranges are adopted from the reference method \cite{gupta2026grasp}. We put the TPE performance threshold on the top 25 percent score range for all optimizers.

\subsection{Hand Optimization}
\Cref{fig:simres}A shows the average sampled hand scores and the best hand score within each iteration. After iteration 200, the score almost converges. We can observe that the optimization curves have a high-frequency noise. Multiple factors may contribute to the noise, such as the stochastic sampling of the hand configurations and the noisy simulation results in Isaac Sim. We also observed that having a higher resolution pad for the palm decreases the noise band with a larger computational load trade-off.

We visualize the three high-, mid-, and low-scored hands that we fabricate in \cref{fig:simres}D. In the lowest-scored designs, the two-finger hands are consistently scored less than the three-finger hands, which is expected in a power grasp task, as fewer contact points and weaker constraints are holding the tool in hand. Parameters such as thumb angle and finger lengths are also different in each hand, yet it is hard to understand the effects of these parameters without a complete parametric analysis, investigating how each design feature affects the final performance score.

\subsection{Parametric Analysis}

Each hand design parameter can influence grasp stability in complex and nonlinear ways. While the optimization process identifies high-performing configurations, it does not directly reveal how individual parameters contribute to performance. To obtain interpretable insights into the design space, we perform a parametric analysis using SHAP (SHapley Additive exPlanations).

We first train a Random Forest regression model to approximate the mapping between hand design parameters and the corresponding grasp stability scores obtained from simulation. The Random Forest model is well-suited for this task due to its ability to capture nonlinear relationships and parameter interactions without requiring explicit functional assumptions. Using SHAP, we compute the contribution of each parameter to the predicted grasp score.

SHAP values provide both global and local interpretability. Globally, they allow us to rank parameters according to their overall importance across the dataset. Locally, they reveal how specific parameter values increase or decrease grasp stability for individual designs. This enables identification of nonlinear effects, interaction patterns, and potential trade-offs between structural and surface-level variables.

By complementing optimization with interpretable analysis, the framework moves beyond black-box search and provides actionable insight into which aspects of morphology most strongly influence grasp performance.

\Cref{fig:simres}B–C present the SHAP analysis results. As shown in Fig.~C, finger number has the strongest contribution to grasp stability. The second most influential parameter is the thumb [$T$] normal offset (directed rightward as illustrated in \cref{fig:Palm}), where larger offsets consistently improve performance. This aligns with physical intuition: increasing the thumb offset enlarges the contact span between the thumb and pinky, enhancing torque resistance, similar to shifting the pinky [$F_2$] outward (blue regions).

In contrast, longer fingers and fingertips consistently degrade performance. Increased link length amplifies required joint torques under external wrenches, reducing grasp robustness. Additionally, elongated links create larger gaps when the hand forms a fist, increasing internal void space and weakening contact enclosure around the object.

Regarding the surface kernels, smaller spread ($\sigma$) and lower maximum deformation generally yield higher grasp scores. The center offsets of the two kernels exhibit opposing effects: a smaller offset for $K_0$ and a larger offset for $K_1$ are associated with improved performance. This indicates that overlapping the kernels, leading to concentrated deformation, is unfavorable for grasp stability.

For clearer visualization of the SHAP results, we group related parameters in \cref{fig:simres}B, stacking them by category and ranking their aggregated contribution to grasp performance.

\subsection{Real-world Dynamic Tasks}

To further evaluate our hand and grasp optimizer and validate the grasp stability metric for tool-use tasks, we conducted human-demonstrated trajectories across three representative tasks on the three fabricated hands mounted on a UR5e robot arm. OptiTrack markers were attached to the tools, and both tool markers and robot end-effector poses were recorded to calculate relative angular motion.

\begin{figure}
    \centering
    \includegraphics[width=\linewidth]{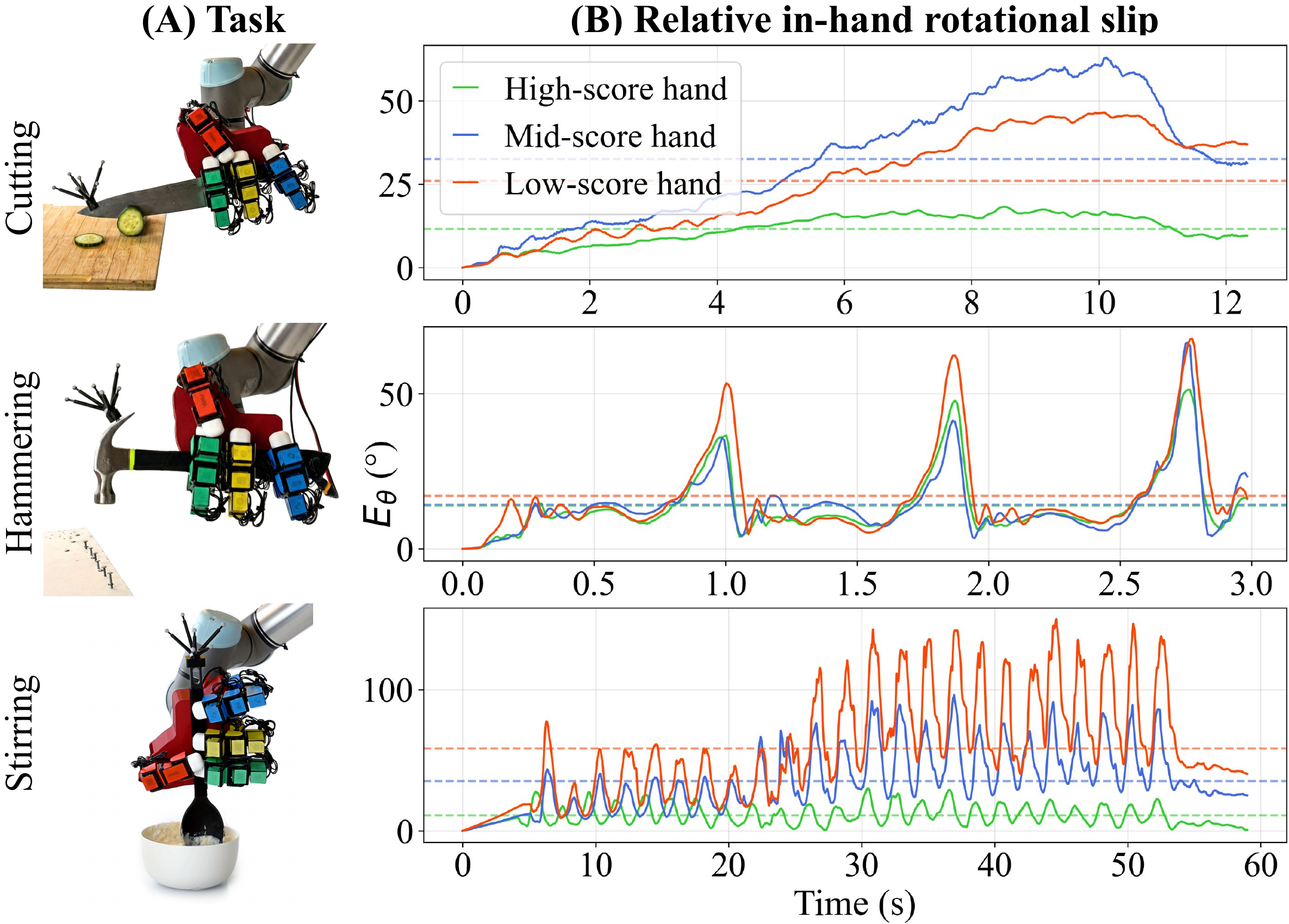}
    \vspace{-18pt}
    \caption{Real-world dynamic task performance. (A) We transfer the optimal grasps obtained in simulation to the real-world test setup and execute the same human-demonstrated trajectories \cite{gupta2026grasp} for three different tasks across three different hand designs, performing five trials for each configuration. (B) In-hand tool rotational slip measured with OptiTrack markers shows tracking errors consistent with simulation-based optimization results. Each curve shows the average error over five trials.}

    \label{fig:expres}
    \vspace{-15pt}
\end{figure}

\Cref{fig:expres} shows the in-hand rotation error, assuming the tool should rigidly follow the end-effector trajectory, for three different hands using their simulation-optimized grasps. Notably, the torque limits used during simulated grasp initialization and the resulting joint targets were directly transferred to the physical hands. We run each hand-tool trajectory for 5 trials and take the average of the results.

The results show that the hand with the highest simulation score consistently outperforms the other two designs in real-world experiments, particularly in the mixing and cutting tasks. In the hammering task, while peak performance is comparable, the high-scoring hand demonstrates more consistent stability over time, exhibiting reduced rotational drift and disturbance sensitivity compared to the other designs.

\begin{table}[t]
\centering
\caption{Optimization parameters in the current power grasp study.}
    \vspace{-5pt}
\label{tab:optimization_parameters}

\begin{tabularx}{\linewidth}{>{\bfseries}l l X c}
\toprule
Category & Opt. Parameter & Range & Count \\
\midrule

Finger pose & Angle [$^\circ$] & [0--30, , -30--0] & 2 \\
 & Normal offset [$mm$]& [0--5, 0--10, 0--5] & 3 \\
 & Side offset [$mm$] & [0--30, , -30--0] & 2 \\

Thumb pose & Angle [$^\circ$] & [-30, 30] & 1 \\
 & Normal offset [$mm$] & [-30, 30] & 1 \\
 & Side offset [$mm$] & [-40, 10] & 1 \\

Palm kernel & Max height  [$mm$] & [0, 20] & 1 \\
 & Spread & [0.05, 0.3] & 2 \\
 & Center angle [$^\circ$] & [0, 360] & 2 \\
 & Center offset & [0, 1] & 2 \\
 & Intensity ratio & [0, 1] & 2 \\

Fingertip & Scale Y & [1.0, 1.5] & 1 \\
 & Scale Z & [0.5, 1.5] & 1 \\

Link lengths & Added length [$mm$] & [0, 10] & 4\\

Structural & Finger code & 1-1-1\ /\ 0-121 & 1 \\
 & Thumb code & 1-22\ /\ 0-22 & 1 \\
 & Finger number & 2\ /\ 3 & 1 \\

\bottomrule

    \vspace{-20pt}
\end{tabularx}
\end{table}

\section{Conclusion} 
\label{sec:conclusion}

We presented a dexterous hand co-design optimization framework with a comprehensive parametric generation stage. The proposed hand generator spans a broad design space, from palm dimensions and finger kinematics to fine-grained surface geometries, enabling systematic, task-driven embodiment optimization.

Hand configurations were evaluated using a simulation-based grasp synthesis and stability assessment pipeline targeting power grasps in tool-use scenarios. To interpret the optimization outcomes, we performed a SHAP-based parametric analysis using a Random Forest surrogate model, revealing the relative importance and interaction effects of structural and surface-level parameters. This analysis provided actionable insight into how morphology influences grasp stability, moving beyond black-box optimization toward interpretable design reasoning.

Real-world experiments following human-demonstrated dynamic trajectories validated the simulation results. The highest-scoring hand in the simulation consistently outperformed alternative designs in mixing and cutting tasks and exhibited more stable behavior over time in hammering. Overall, the results demonstrate that our framework enables interpretable, task-specific morphological optimization with reliable sim-to-real transfer.

The framework can generalize to other hand designs and tasks. By redefining the parameterized design space, the same pipeline can optimize different robotic hands for different objectives.

\bibliographystyle{IEEEtran}
\bibliography{IEEEabrv, references}

\begin{thebibliography}{10}
\providecommand{\url}[1]{#1}
\csname url@samestyle\endcsname
\providecommand{\newblock}{\relax}
\providecommand{\bibinfo}[2]{#2}
\providecommand{\BIBentrySTDinterwordspacing}{\spaceskip=0pt\relax}
\providecommand{\BIBentryALTinterwordstretchfactor}{4}
\providecommand{\BIBentryALTinterwordspacing}{\spaceskip=\fontdimen2\font plus
\BIBentryALTinterwordstretchfactor\fontdimen3\font minus \fontdimen4\font\relax}
\providecommand{\BIBforeignlanguage}[2]{{%
\expandafter\ifx\csname l@#1\endcsname\relax
\typeout{** WARNING: IEEEtran.bst: No hyphenation pattern has been}%
\typeout{** loaded for the language `#1'. Using the pattern for}%
\typeout{** the default language instead.}%
\else
\language=\csname l@#1\endcsname
\fi
#2}}
\providecommand{\BIBdecl}{\relax}
\BIBdecl

\bibitem{gilday2023sensing}
K.~Gilday, J.~Hughes, and F.~Iida, ``Sensing, actuating, and interacting through passive body dynamics: A framework for soft robotic hand design,'' \emph{Soft Robotics}, vol.~10, no.~1, pp. 159--173, 2023.

\bibitem{marzke2000evolution}
M.~W. Marzke and R.~F. Marzke, ``Evolution of the human hand: approaches to acquiring, analysing and interpreting the anatomical evidence,'' \emph{The Journal of Anatomy}, vol. 197, no.~1, pp. 121--140, 2000.

\bibitem{christoph2025orca}
C.~C. Christoph, M.~Eberlein, F.~Katsimalis, A.~Roberti, A.~Sympetheros, M.~R. Vogt, D.~Liconti, C.~Yang, B.~G. Cangan, R.~J. Hinchet \emph{et~al.}, ``Orca: An open-source, reliable, cost-effective, anthropomorphic robotic hand for uninterrupted dexterous task learning,'' in \emph{2025 IEEE/RSJ International Conference on Intelligent Robots and Systems (IROS)}.\hskip 1em plus 0.5em minus 0.4em\relax IEEE, 2025, pp. 8503--8510.

\bibitem{shaw2023leap}
K.~Shaw, A.~Agarwal, and D.~Pathak, ``Leap hand: Low-cost, efficient, and anthropomorphic hand for robot learning,'' \emph{arXiv preprint arXiv:2309.06440}, 2023.

\bibitem{calandra2018more}
R.~Calandra, A.~Owens, D.~Jayaraman, J.~Lin, W.~Yuan, J.~Malik, E.~H. Adelson, and S.~Levine, ``More than a feeling: Learning to grasp and regrasp using vision and touch,'' \emph{IEEE Robotics and Automation Letters}, vol.~3, no.~4, pp. 3300--3307, 2018.

\bibitem{yin2025geometric}
Z.-H. Yin, C.~Wang, L.~Pineda, K.~Bodduluri, T.~Wu, P.~Abbeel, and M.~Mukadam, ``Geometric retargeting: A principled, ultrafast neural hand retargeting algorithm,'' in \emph{2025 IEEE/RSJ International Conference on Intelligent Robots and Systems (IROS)}.\hskip 1em plus 0.5em minus 0.4em\relax IEEE, 2025, pp. 17\,376--17\,382.

\bibitem{cheney2014unshackling}
N.~Cheney, R.~MacCurdy, J.~Clune, and H.~Lipson, ``Unshackling evolution: evolving soft robots with multiple materials and a powerful generative encoding,'' \emph{ACM SIGEVOlution}, vol.~7, no.~1, pp. 11--23, 2014.

\bibitem{bai2025learning}
X.~Bai, N.~Hansen, A.~Singh, M.~T. Tolley, Y.~Duan, P.~Abbeel, X.~Wang, and S.~Yi, ``Learning to design soft hands using reward models,'' \emph{arXiv preprint arXiv:2510.17086}, 2025.

\bibitem{pathak2019learning}
D.~Pathak, C.~Lu, T.~Darrell, P.~Isola, and A.~A. Efros, ``Learning to control self-assembling morphologies: a study of generalization via modularity,'' \emph{Advances in Neural Information Processing Systems}, vol.~32, 2019.

\bibitem{deimel2017automated}
R.~Deimel, P.~Irmisch, V.~Wall, and O.~Brock, ``Automated co-design of soft hand morphology and control strategy for grasping,'' in \emph{2017 IEEE/RSJ International Conference on Intelligent Robots and Systems (IROS)}.\hskip 1em plus 0.5em minus 0.4em\relax IEEE, 2017, pp. 1213--1218.

\bibitem{fay2025cross}
K.~Fay, D.~A. Djapri, A.~Zorin, J.~Clinton, A.~E. Lahib, H.~Su, M.~T. Tolley, S.~Yi, and X.~Wang, ``Cross-embodied co-design for dexterous hands,'' \emph{arXiv preprint arXiv:2512.03743}, 2025.

\bibitem{yi2025co}
S.~Yi, X.~Bai, A.~Singh, J.~Ye, M.~T. Tolley, and X.~Wang, ``Co-design of soft gripper with neural physics,'' \emph{arXiv preprint arXiv:2505.20404}, 2025.

\bibitem{mannam2023framework}
P.~Mannam, K.~Shaw, D.~Bauer, J.~Oh, D.~Pathak, and N.~Pollard, ``A framework for designing anthropomorphic soft hands through interaction,'' \emph{arXiv preprint arXiv:2306.04784}, 2023.

\bibitem{meixner2019automated}
A.~Meixner, C.~Hazard, and N.~Pollard, ``Automated design of simple and robust manipulators for dexterous in-hand manipulation tasks using evolutionary strategies,'' in \emph{2019 IEEE-RAS 19th International Conference on Humanoid Robots (Humanoids)}.\hskip 1em plus 0.5em minus 0.4em\relax IEEE, 2019, pp. 281--288.

\bibitem{mannam2024design}
P.~Mannam, X.~Liu, D.~Zhao, J.~Oh, and N.~Pollard, ``Design and control co-optimization for automated design iteration of dexterous anthropomorphic soft robotic hands,'' in \emph{2024 IEEE 7th International Conference on Soft Robotics (RoboSoft)}.\hskip 1em plus 0.5em minus 0.4em\relax IEEE, 2024, pp. 332--339.

\bibitem{vaish2024co}
A.~Vaish and O.~Brock, ``Co-designing manipulation systems using task-relevant constraints,'' in \emph{2024 IEEE International Conference on Robotics and Automation (ICRA)}.\hskip 1em plus 0.5em minus 0.4em\relax IEEE, 2024, pp. 4177--4183.

\bibitem{you2019kinematic}
W.~S. You, Y.~H. Lee, G.~Kang, H.~S. Oh, J.~K. Seo, and H.~R. Choi, ``Kinematic design optimization for anthropomorphic robot hand based on interactivity of fingers,'' \emph{Intelligent Service Robotics}, vol.~12, no.~2, pp. 197--208, 2019.

\bibitem{weisz2012pose}
J.~Weisz and P.~K. Allen, ``Pose error robust grasping from contact wrench space metrics,'' in \emph{2012 IEEE international conference on robotics and automation}.\hskip 1em plus 0.5em minus 0.4em\relax IEEE, 2012, pp. 557--562.

\bibitem{miller2004graspit}
A.~T. Miller and P.~K. Allen, ``Graspit! a versatile simulator for robotic grasping,'' \emph{IEEE Robotics \& Automation Magazine}, vol.~11, no.~4, pp. 110--122, 2004.

\bibitem{wang2022dexgraspnet}
R.~Wang, J.~Zhang, J.~Chen, Y.~Xu, P.~Li, T.~Liu, and H.~Wang, ``Dexgraspnet: A large-scale robotic dexterous grasp dataset for general objects based on simulation,'' \emph{arXiv preprint arXiv:2210.02697}, 2022.

\bibitem{luck2020data}
K.~S. Luck, H.~B. Amor, and R.~Calandra, ``Data-efficient co-adaptation of morphology and behaviour with deep reinforcement learning,'' in \emph{Conference on Robot Learning}.\hskip 1em plus 0.5em minus 0.4em\relax PMLR, 2020, pp. 854--869.

\bibitem{rios2013derivative}
L.~M. Rios and N.~V. Sahinidis, ``Derivative-free optimization: a review of algorithms and comparison of software implementations,'' \emph{Journal of Global Optimization}, vol.~56, no.~3, pp. 1247--1293, 2013.

\bibitem{larson2019derivative}
J.~Larson, M.~Menickelly, and S.~M. Wild, ``Derivative-free optimization methods,'' \emph{Acta Numerica}, vol.~28, pp. 287--404, 2019.

\bibitem{grebenstein2012hand}
M.~Grebenstein, M.~Chalon, W.~Friedl, S.~Haddadin, T.~Wimb{\"o}ck, G.~Hirzinger, and R.~Siegwart, ``The hand of the dlr hand arm system: Designed for interaction,'' \emph{The International Journal of Robotics Research}, vol.~31, no.~13, pp. 1531--1555, 2012.

\bibitem{cabas2006optimized}
R.~Cab{\'a}s, L.~M. Cabas, and C.~Balaguer, ``Optimized design of the underactuated robotic hand,'' in \emph{Proceedings 2006 IEEE International Conference on Robotics and Automation, 2006. ICRA 2006.}\hskip 1em plus 0.5em minus 0.4em\relax IEEE, 2006, pp. 982--987.

\bibitem{dong2018geometric}
H.~Dong, E.~Asadi, C.~Qiu, J.~Dai, and I.-M. Chen, ``Geometric design optimization of an under-actuated tendon-driven robotic gripper,'' \emph{Robotics and Computer-Integrated Manufacturing}, vol.~50, pp. 80--89, 2018.

\bibitem{xu2021end}
J.~Xu, T.~Chen, L.~Zlokapa, M.~Foshey, W.~Matusik, S.~Sueda, and P.~Agrawal, ``An end-to-end differentiable framework for contact-aware robot design,'' \emph{arXiv preprint arXiv:2107.07501}, 2021.

\bibitem{fayhardware}
K.~Fay, S.~Yi, M.~T. Tolley, X.~Wang, and H.~Su, ``Hardware optimization for in-hand rotation,'' in \emph{1st Workshop on Robot Hardware-Aware Intelligence}, 2025.

\bibitem{gilday2025embodied}
K.~Gilday, C.~Sirithunge, F.~Iida, and J.~Hughes, ``Embodied manipulation with past and future morphologies through an open parametric hand design,'' \emph{Science Robotics}, vol.~10, no. 102, p. eads6437, 2025.

\bibitem{ye2025power}
J.~Ye, L.~Wei, G.~Jiang, C.~Jing, X.~Zou, and X.~Wang, ``From power to precision: Learning fine-grained dexterity for multi-fingered robotic hands,'' \emph{arXiv preprint arXiv:2511.13710}, 2025.

\bibitem{yu1998optimization}
Y.~Yu, K.~Takeuchi, and T.~Yoshikawa, ``Optimization of robot hand power grasps,'' in \emph{Proceedings. 1998 IEEE International Conference on Robotics and Automation (Cat. No. 98CH36146)}, vol.~4.\hskip 1em plus 0.5em minus 0.4em\relax IEEE, 1998, pp. 3341--3347.

\bibitem{mayorga1992kinematic}
R.~V. Mayorga, B.~Ressa, and A.~K. Wong, ``A kinematic design optimization of robot manipulators,'' in \emph{Proceedings 1992 IEEE International Conference on Robotics and Automation}.\hskip 1em plus 0.5em minus 0.4em\relax IEEE, 1992, pp. 396--401.

\bibitem{vijaykumar1985geometric}
R.~Vijaykumar, M.~Tsai, and K.~Waldron, ``Geometric optimization of manipulator structures for working volume and dexterity,'' in \emph{Proceedings. 1985 IEEE International Conference on Robotics and Automation}, vol.~2.\hskip 1em plus 0.5em minus 0.4em\relax IEEE, 1985, pp. 228--236.

\bibitem{ha2017joint}
S.~Ha, S.~Coros, A.~Alspach, J.~Kim, and K.~Yamane, ``Joint optimization of robot design and motion parameters using the implicit function theorem.'' in \emph{Robotics: Science and systems}, vol.~13, 2017, pp. 10--15\,607.

\bibitem{chen2023c}
C.~Chen, P.~Xiang, H.~Lu, Y.~Wang, and R.~Xiong, ``C 2: Co-design of robots via concurrent-network coupling online and offline reinforcement learning,'' in \emph{2023 IEEE/RSJ International Conference on Intelligent Robots and Systems (IROS)}.\hskip 1em plus 0.5em minus 0.4em\relax IEEE, 2023, pp. 7487--7494.

\bibitem{chen2021co}
T.~Chen, Z.~He, and M.~Ciocarlie, ``Co-designing hardware and control for robot hands,'' \emph{Science Robotics}, vol.~6, no.~54, p. eabg2133, 2021.

\bibitem{whitman2020modular}
J.~Whitman, M.~Travers, and H.~Choset, ``Modular mobile robot design selection with deep reinforcement learning,'' in \emph{NeurIPS Workshop on ML for engineering modeling, simulation and design}, vol.~2, 2020, pp. 7--2.

\bibitem{whitman2023learning}
------, ``Learning modular robot control policies,'' \emph{IEEE Transactions on Robotics}, vol.~39, no.~5, pp. 4095--4113, 2023.

\bibitem{schaff2019jointly}
C.~Schaff, D.~Yunis, A.~Chakrabarti, and M.~R. Walter, ``Jointly learning to construct and control agents using deep reinforcement learning,'' in \emph{2019 international conference on robotics and automation (ICRA)}.\hskip 1em plus 0.5em minus 0.4em\relax IEEE, 2019, pp. 9798--9805.

\bibitem{gupta2026grasp}
H.~Gupta, M.~A. Mirzaee, and W.~Yuan, ``Grasp to act: Dexterous grasping for tool use in dynamic settings,'' \emph{arXiv preprint arXiv:2602.20466}, 2026.

\end{thebibliography}

\end{document}